\documentclass[10pt,twocolumn,letterpaper]{article}

\usepackage[pagenumbers]{wacv} 

\usepackage{graphicx}
\usepackage{amsmath}
\usepackage{amssymb}
\usepackage{booktabs}

\usepackage{multirow}
\usepackage{color,soul}

\usepackage[pagebackref,breaklinks,colorlinks]{hyperref}

\usepackage[capitalize]{cleveref}
\crefname{section}{Sec.}{Secs.}
\Crefname{section}{Section}{Sections}
\Crefname{table}{Table}{Tables}
\crefname{table}{Tab.}{Tabs.}

\def\wacvPaperID{2224} 
\def\confName{WACV}
\def\confYear{2025}

\begin{document}

\title{LLS: \underline{L}ocal \underline{L}earning Rule for Deep Neural Networks Inspired by Neural Activity \underline{S}ynchronization}

\author{Marco~P.~E.~Apolinario,\quad Arani~Roy,\quad Kaushik~Roy\\
Elmore Family School of Electrical and Computer Engineering\\ Purdue University, West Lafayette, IN, USA \\
{\tt\small \{mapolina, roy173, kaushik\}@purdue.edu}
}
\maketitle

\begin{abstract}
   Training deep neural networks (DNNs) using traditional backpropagation (BP) presents challenges in terms of computational complexity and energy consumption, particularly for on-device learning where computational resources are limited. 
  Various alternatives to BP, including random feedback alignment, forward-forward, and local classifiers, have been explored to address these challenges. 
  These methods have their advantages, but they can encounter difficulties when dealing with intricate visual tasks or demand considerable computational resources. 
  In this paper, we propose a novel Local Learning rule inspired by neural activity Synchronization phenomena (LLS) observed in the brain. LLS utilizes fixed periodic basis vectors to synchronize neuron activity within each layer, enabling efficient training without the need for additional trainable parameters. 
  We demonstrate the effectiveness of LLS and its variations, LLS-M and LLS-MxM, on multiple image classification datasets, achieving accuracy comparable to BP with reduced computational complexity and minimal additional parameters.
  Specifically, LLS achieves comparable performance with up to $300 \times$ fewer multiply-accumulate (MAC) operations and half the memory requirements of BP.
  Furthermore, the  performance of LLS on the Visual Wake Word (VWW) dataset highlights its suitability for on-device learning tasks, making it a promising candidate for edge hardware implementations. 
  Our code is available at \href{https://github.com/mapolinario94/LLS-DNN}{GitHub repository}.
\end{abstract}

\section{Introduction}\label{sec:introduction}

Currently, stochastic gradient-based optimization schemes serve as the default method for training deep neural network (DNN) models. 
These schemes leverage the backpropagation (BP) algorithm, enabling the computation of gradients of the loss function with respect to the trainable parameters (weights) in the hidden layers. 
However, BP is associated with high time and memory complexities, leading to significant energy consumption. 
For instance, in a model with $L$ layers and $n$ neurons per layer, BP exhibits time and memory complexities of $O(Ln^2)$ and $O(Ln)$, respectively. 
While suitable for offline training in environments with ample computational resources (such as the cloud), these computational demands render BP inefficient for on-device learning on low-power edge devices, where computation resources are severely constrained \cite{Zhang2018SignSystems,  Ankit2020PANTHER:ReRAM, Peng2021NeuroSimV2}. 
Studies such as \cite{Ankit2020PANTHER:ReRAM} and \cite{Peng2021NeuroSimV2} highlight the large energy consumption associated with extensive external memory accesses and gradient computations in BP.
Consequently, there is a need for hardware-friendly algorithms to facilitate efficient on-device learning on low-power edge devices.

With this consideration in mind, numerous works have explored alternatives to backpropagation (BP), trying to eliminate the need of computationally expensive gradient calculations associated with BP. 
Methods like feedback alignment (FA) and its variant, direct feedback alignment (DFA), utilize random matrices to propagate error signals or directly project errors to each layer, offering some reduction in dependency across layers but still requiring similar memory demands \cite{Lillicrap2016RandomLearning, Trondheim2016DirectNetworks, Crafton2019DirectLearning}.
An alternative to this approach is proposed by \cite{Frenkel2021LearningNetworks}, which uses random matrices to project targets instead of errors, thereby enabling each layer to be updated independently. 
Although promising, these methods do not scale well for deep neural networks (DNNs). 
In contrast, \cite{Ororbia2023Backpropagation-FreeAlignment} proposes a local learning rule that matches BP performance in large models at the cost of significantly increasing the number of trainable parameters and computational complexity. 
Recent research works have attempted to replace BP’s backward pass with an additional forward pass, aiming to enhance biological plausibility, though they suffer from slow convergence and have not yet proven effective for deep networks \cite{Dellaferrera2022Error-drivenPass, Hinton2022TheInvestigations}. 
Additionally, \cite{Journe2023HebbianFeedback} proposes a biologically inspired method using a soft winner-take-all mechanism to facilitate unsupervised learning in simpler DNN models.
In contrast, \cite{Nkland2019TrainingSignals, Belilovsky2018GreedyImageNet} and \cite{Wang2021RevisitingTraining} proposed to use auxiliary networks as local classifiers. 
These methods \cite{Nkland2019TrainingSignals, Belilovsky2018GreedyImageNet, Wang2021RevisitingTraining} avoid using end-to-end BP by breaking the problem into smaller pieces and generating error signals with the aid of such local classifiers per layer or group of layers.
Since these methods necessitate additional layers to generate the learning signal, we categorize them as hybrids between local learning and BP.

The aforementioned learning methods often struggle to scale to complex vision tasks without high computational costs \cite{Lillicrap2016RandomLearning, Trondheim2016DirectNetworks, Frenkel2021LearningNetworks, Dellaferrera2022Error-drivenPass, Journe2023HebbianFeedback, Ororbia2023Backpropagation-FreeAlignment}. 
Hybrid approaches using local classifiers \cite{Nkland2019TrainingSignals, Belilovsky2018GreedyImageNet} offer a better balance for on-device learning but at the cost of increasing trainable parameters, thus increasing memory and energy demands. 
To address this, we propose a Local Learning rule inspired by brain-like neural activity Synchronization (LLS). 
This rule bypasses intensive gradient calculations of BP and scales to complex vision tasks and deep networks. 

Neuronal activity synchronization in the brain reflects the correlation of brain signals. 
Studies in \cite{Jutras2010SynchronousFormation, GuevaraErra2017NeuralDynamics, Miller2014VisualEnsembles, Hopfield1982NeuralAbilities., Carrillo-Reid2019ControllingEnsembles}, have demonstrated that neuronal ensembles in the brain synchronize their activity during cognitive learning processes or in response to visual stimuli. 
Inspired from this biological process, LLS utilizes fixed periodic basis vectors to synchronize neuron activity within same layers of the model.
Our experiments show that simple periodic functions like cosine and square enable effective learning in complex image classification tasks. 
These functions are computationally lightweight, allowing on-the-fly generation on low-power devices without additional trainable parameters. 
Furthermore, we explore variations of LLS, such as LLS-M and LLS-MxM, to enhance performance on more complex tasks. 
LLS-M learns to modulate the amplitude of the fixed basis, while LLS-MxM learns to construct an improved basis through a linear combination of the fixed basis.
Both variants require minimal trainable parameters, on the order of $O(C)$ and $O(C^2)$, where $C$ represents the number of classes. 
Evaluation on public image classification datasets, including CIFAR10, CIFAR100, IMAGENETTE, TinyIMAGENET, and Visual Wake Words (VWW), demonstrates that our method achieves high accuracy comparable to BP, with significant reductions in MAC operations, memory usage, and minimal additional parameters. 
Notably, our method's performance on the VWW dataset underscores its suitability for on-device learning hardware implementations.

The main contributions of the paper are as follows:
\begin{itemize}
    \item A novel local learning rule that utilizes fixed periodic basis vectors to synchronize neural activity per layer, achieving high accuracy with reduced MAC operations, memory usage, and minimal additional trainable parameters.
    \item Evaluation of the effectiveness of our method on various image classification datasets, demonstrating accuracy comparable to BP.
    \item Demonstration of the suitability of our method for on-device learning tasks by evaluating its performance on the Visual Wake Word (VWW) dataset, achieving high performance with low computational complexity.
\end{itemize}

\section{Background}\label{sec:background}
    \subsection{Backpropagation (BP)}
    As noted earlier, the backpropagation (BP) algorithm is central to deep learning. 
    We explore its mechanics here and introduce key notations used in this work. 
    A neural network model can be represented as a parameterized function $F(\mathbf{x} ; \mathbf{\theta})$, where $\mathbf{x}$ is the input data and $\mathbf{\theta}$ are the parameters. 
    For an $L$-layer model, the parameters are $\mathbf{\theta}=[\mathbf{w}^{(1)}, \cdots, \mathbf{w}^{(L)}]$, with $\mathbf{w}^{(l)}$ representing the weights of the $l$-th layer. 
    Each layer produces an output, $\mathbf{h}^{(l)}$, obtained by applying a linear transformation over the input $\mathbf{h}^{(l-1)}$ based on the parameters $\mathbf{w}^{(l)}$, resulting in an intermediate representation $\mathbf{z}^{(l)}$, followed by a non-linear element-wise activation function $\mathbf{h}^{(l)}=f(\mathbf{z}^{(l)})$. 
    Given a loss function $\mathcal{L}$ and a labeled dataset $[\mathbf{X}, \mathbf{Y}^*]$, where $\mathbf{X}$ are the inputs and $\mathbf{Y}^*$ the labels. 
    The objective is to find the parameters $\mathbf{\theta}$ that minimize the loss, i.e., $\mathbf{\theta}:=\arg\min_{\mathbf{\theta}} \mathcal{L}(\mathbf{Y}^*, F(\mathbf{X}; \mathbf{\theta}))$. 
    For this purpose, the conventional approach is to use mini-batch stochastic gradient descent (SGD), which randomly samples a mini-batch of data $[\mathbf{x}, \mathbf{y}^*]$ from the dataset to estimate the gradient of the loss function. 
    Such a learning algorithm, with a learning rate ($\eta$), has the following update rule for the parameters:
    \begin{equation}
        \mathbf{w}^{(l)} := \mathbf{w}^{(l)} - \eta \nabla_{\mathbf{w}^{(l)}} \mathcal{L} 
        \label{eq:sgd}
    \end{equation}
    The gradient $\nabla_{\mathbf{w}^{(l)}} \mathcal{L}$ is computed based on the BP algorithm. 
    BP operates in two phases: the forward pass and the backward pass. 
    During the forward pass, an input $\mathbf{x}$ is propagated layer by layer through the model to obtain a model prediction $\mathbf{h}^{(L)}=F(\mathbf{x}; \mathbf{\theta})$, and the loss $\mathcal{L}(\mathbf{y}^*, \mathbf{h}^{(L)})$ is computed. 
    In this process, all intermediate representations $\mathbf{z}^{(l)}$ are saved. 
    Then, in the backward pass, the chain rule is used to compute the gradients as follows:
    \begin{equation}\label{eq:bp_rule}
    \begin{split}
        \nabla_{\mathbf{w}^{(l)}} \mathcal{L}&=\frac{\partial\mathcal{L}}{\partial\mathbf{h}^{(l)}}\frac{\partial \mathbf{h}^{(l)}}{\partial \mathbf{z}^{(l)}}\frac{\partial \mathbf{z}^{(l)}}{\partial \mathbf{w}^{(l)}}\\ &=\frac{\partial\mathcal{L}}{\partial\mathbf{h}^{(L)}} \prod^{L}_{i=l+1}\frac{\partial \mathbf{h}^{(i)}}{\partial \mathbf{h}^{(i-1)}}\frac{\partial \mathbf{h}^{(l)}}{\partial \mathbf{z}^{(l)}}\frac{\partial \mathbf{z}^{(l)}}{\partial \mathbf{w}^{(l)}} 
    \end{split}
    \end{equation}
    Here, $\frac{\partial\mathcal{L}}{\partial\mathbf{h}^{(l)}}$ is the learning signal obtained by propagating errors from the last layer ($L$) to layer $l$. 
    Additionally, $\frac{\partial \mathbf{h}^{(l)}}{\partial \mathbf{z}^{(l)}}$ corresponds to the derivative of the activation function $f'(\mathbf{z}^{(l)})$, and $\frac{\partial \mathbf{z}^{(l)}}{\partial \mathbf{w}^{(l)}}$ is equivalent to the input of the $l$-th layer, i.e., $\mathbf{h}^{(l-1)}$. 
    From (\ref{eq:bp_rule}), it can be observed that while the latter two factors on the right-hand side of (\ref{eq:bp_rule}) depend only on the inputs and outputs of layer $l$, the learning signal depends on all successive layers. 
    Therefore, 
    the weight updates must be sequential (i.e., update-locking problem). 
    Moreover, the computational and memory complexity of BP are $O(Ln^2)$ and $O(Ln)$, respectively, with $n$ representing the average number of neurons per layer.

    \subsection{Local learning for DNN}\label{sec:local_learning}
    The non-locality and update-locking features of BP, among others, have been argued as reasons that make BP unlikely as the learning rule used by the brain \cite{Lillicrap2020BackpropagationBrain}. 
    Different local learning mechanisms that may not rely on the propagation of errors using symmetric weights have been explored in many works \cite{Trondheim2016DirectNetworks, Frenkel2021LearningNetworks, Dellaferrera2022Error-drivenPass, Hinton2022TheInvestigations, Journe2023HebbianFeedback}. 
    Here, we refer to local learning as learning rules that compute weight updates ($\Delta \mathbf{w}^{(l)}$) based only on inputs ($\mathbf{h}^{(l-1)}$), outputs ($\mathbf{z}^{(l)}$), and some other global factors. 
    An example is the DFA method \cite{Trondheim2016DirectNetworks}, which uses random feedback weights ($\mathbf{B}^{(l)}$) to produce the learning signal. 
    In this method, $\frac{\partial\mathcal{L}}{\partial\mathbf{h}^{(l)}}$ in (\ref{eq:bp_rule}) is replaced by $\frac{\partial\mathcal{L}}{\partial\mathbf{h}^{(L)}}\mathbf{B}^{(l)}$. 
    A similar method is proposed by \cite{Frenkel2021LearningNetworks}, denoted as DRTP, which uses fixed random learning signals produced by propagating the labels instead of error. 
    In other words, the learning signals are $\mathbf{y}^*\mathbf{B}^{(l)}$. 
    Other approaches, such as those by \cite{Dellaferrera2022Error-drivenPass, Hinton2022TheInvestigations}, use two forward passes to produce the learning signal, or produce a learning signal based on a soft competition mechanism as proposed by \cite{Journe2023HebbianFeedback}.

    \subsection{Neural activity synchronization in the brain}\label{sec:neural_sync}
    Neural activity synchronization refers to the correlated neuronal signals across different regions of the brain. 
    Groups of neurons that co-activate in response to sensory stimuli or during spontaneous activity are often referred to as ensembles. 
    These ensembles play a crucial role in various cognitive functions, including the processing of visual stimuli in the cortex \cite{Miller2014VisualEnsembles}, memory formation \cite{Hopfield1982NeuralAbilities.}, and behavior regulation \cite{Carrillo-Reid2019ControllingEnsembles}. 
    In addition to these roles, modulations in oscillatory neuronal activity are commonly observed when humans engage in cognitive tasks. 
    For instance, as highlighted by \cite{GuevaraErra2017NeuralDynamics}, the complex, high-dimensional dynamics of neuronal activity can collapse into low-dimensional oscillatory modes, which in turn facilitates memory enhancement and learning. 
    This synchronization not only simplifies the representation of neuronal dynamics but also captures both linear and non-linear aspects of neuronal interactions. 
    Drawing inspiration from these biological processes, we propose a local learning rule (LLS) that employs fixed periodic vectors for each class to synchronize neural activity within the same layer of a neural network. 
    This approach is intended to enhance the efficiency of learning in artificial systems. 
    By using periodic vectors, the LLS encourages groups of neurons, distributed periodically within the same layer, to exhibit high activity in response to specific visual stimuli (such as images of a particular class). 
    This design is inspired in the concept of neuronal ensembles within artificial neural networks.


\section{LLS: Local Learning Rule inspired by Neural Activity Synchronization}\label{sec:lls}

    \begin{figure*}
    \centering
    \includegraphics[width=\textwidth]{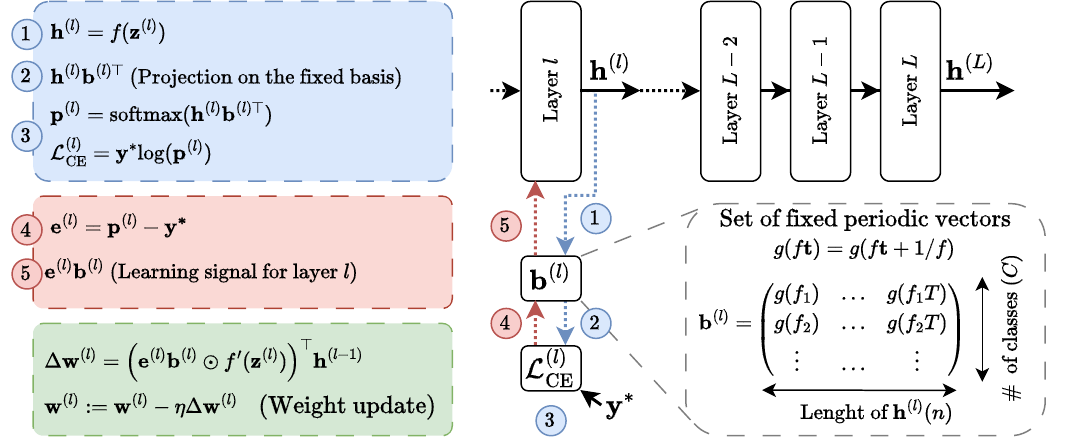}
    \caption{Overview of LLS. Weight updates for the $l$-th hidden layer within an $L$-layer neural network are derived via per-layer minimization of cross-entropy loss ($\mathcal{L}^{(l)}{\mathrm{CE}}$) on the projection of output activations ($\mathbf{h}^{(l)}$) over a fixed basis of periodic vectors $\mathbf{b}^{(l)}$, i.e., $\mathbf{h}^{(l)}\mathbf{b}^\top$. This produces a local error signal as the difference between the softmax of the projection ($\mathbf{p}^{(l)}$) and the one-hot encoded labels $\mathbf{y}^*$. Subsequently, this error signal is multiplied with the fixed basis to generate the learning signal. Weight updates are then determined by multiplying the locally generated learning signal with the layer's inputs and outputs. Consequently, LLS enables independent layer updates based on local information, resulting in low time and memory complexities of $O(LCn)$ and $O(n_{max})$, respectively. It is noteworthy that the fixed basis $\mathbf{b}^{(l)}$ comprises $C$ vectors, where $C$ represents the number of classes for the classification task. Furthermore, the fixed basis vectors are constructed using periodic functions $g(f_c, t) = g(f_c, t + 1/f_c)$, where $f_c$ denotes the spatial frequency associated with class $c$. }
    \label{fig:method_diagram}
    \end{figure*}

    LLS aims to synchronize neural activity within the same layer while minimizing computational complexity and additional trainable parameters. 
    We emphasize three core aspects of LLS: (1) locality, (2) update-unlocking, and (3) minimal parameter requirements.
    
    First, LLS operates locally within each layer, updating synaptic connections ($\mathbf{w}^{(l)}$) based on local inputs ($\mathbf{h}^{(l-1)}$), outputs ($\mathbf{h}^{(l)}$), and generated learning signals. 
    The locally generated learning signals are obtained by projecting $\mathbf{h}^{(l)}$ onto a set of fixed periodic basis vectors $\mathbf{b}^{(l)}$, which align with specific classes to optimize layer performance. 
    Local operation reduces computational overhead of computing the weight gradients.
    
    Second, LLS's update-unlocking feature is a by-product of locality and  enables independent weight updates per layer, eliminating the need to save the output activations of all the layers in the model during training. 
    This results in a memory complexity of $O(n_{max})$, where $n_{max}$ represents the maximum number of neurons in a layer. 
    Unlike methods employing auxiliary local classifiers, LLS requires no additional trainable parameters, utilizing fixed periodic vectors for alignment. 
    However, for tasks with numerous classes, relying solely on fixed vectors may present challenges, as discussed in Section~\ref{sec:experimental_evaluation}.
    To address these limitations, we also propose LLS-M and LLS-MxM as variations of LLS. 
    LLS-M enables learning of optimal modulation for fixed basis vectors, while LLS-MxM learns to form a superior basis via a linear combination of fixed vectors. 
    Both variations entail minimal additional trainable parameters on the order of $O(C)$ and $O(C^2)$, respectively, where $C$ denotes the number of classes in a task.

    \subsection{Technical details}\label{sec:technical_lls}
    The hidden layers are trained based on the alignment of their output activations ($\mathbf{h}^{(l)}$) with predefined set of fixed basis vectors ($\mathbf{b}^{(l)}$), as shown in Fig.~\ref{fig:method_diagram}.
    Alignment is measured as the inner product of a layer's output activations and the basis.
    To encourage synchronicity in neural responses among neurons, the fixed basis vectors are constructed using periodic functions $g(f, t) = g(f, t + 1/f)$, where $f$ represents spatial frequency.
    
    For a classification problem with $C$ classes, each class $c$ has its own vector $\mathbf{b}^{(l)}_c=g(f_c, \mathbf{t}^{(l)})$ where $\mathbf{t}^{(l)}=[1,2,3, \cdots, T^{(l)}]$, $T^{(l)} = $ the length of $l$-th layer's output ($\mathbf{h}^{(l)}$) and $f_c = $ a fixed frequency for class $c$.
    Note that these basis vectors have the same frequencies for all layers but with different lengths.
    The weight updates can be derived as a per-layer minimization of cross-entropy loss ($\mathcal{L}^{(l)}$) on the projection of the activations over the fixed basis ($\mathbf{h}^{(l)}\mathbf{b}^\top$), as illustrated in Fig.~\ref{fig:method_diagram}.
    Specifically, the per-layer cross-entropy loss is described as follows:
    \begin{equation}
    \begin{split}
        \mathcal{L}^{(l)}(\mathbf{h}^{(l)}, \mathbf{y^*}) &= -\frac{1}{N}\sum^N_{n=1} \mathbf{y^*}\textrm{log}(\mathbf{p}^{(l)}_{n}) \\ & = -\frac{1}{N}\sum^N_{n=1} \textrm{log}\frac{\textrm{exp}(\mathbf{h}^{(l)}_n\mathbf{b}^\top_{c_n^*})}{\sum_{c=1}^C \textrm{exp}(\mathbf{h}^{(l)}_n\mathbf{b}^\top_{c})}
    \end{split}
    \end{equation}
    Here, $N$ is the number of samples in the mini-batch, $c^{*}_n$ is the class index for the $n$-th sample in the mini-batch, and $\mathbf{p}^{(l)}_n$ is probability vector obtained of applying the softmax function over the projection vector $\mathbf{h}^{(l)}_n\mathbf{b}^\top$.
    Solving the per-layer minimization problem, $\min_{w^{(l)}}\mathcal{L}^{(l)}(\mathbf{h}^{(l)}, \mathbf{y^*})$, results in the following expression for weight updates on the $l$-th layer:
    \begin{equation}
    \begin{split}
        \Delta \mathbf{w}^{(l)} &= \frac{1}{N}\left((\mathbf{p}^{(l)} - \mathbf{y^*})\mathbf{b}^{(l)}\odot f'(\mathbf{z}^{(l)})\right)^\top  \mathbf{h}^{(l-1)} \\
        & = \frac{1}{N}\left(\mathbf{e}^{(l)}\mathbf{b}^{(l)}\odot f'(\mathbf{z}^{(l)})\right)^\top  \mathbf{h}^{(l-1)}
        \label{eq:lls_update}
    \end{split}
    \end{equation}
    From Equation (\ref{eq:lls_update}), it is evident that the weight updates for each layer $l$ depend solely on the local variables of that layer, including its inputs, outputs, and the set of fixed basis vectors.
    Consequently, all layers can be updated independently of the rest of the model. 
    These independent updates are the reason why the memory complexity of LLS depends only on the largest layer (the layer with the highest number of neurons), in contrast with end-to-end training methods that require memory proportional to the number of neurons in the entire model.
    Moreover, since LLS's learning signals are generated locally, the time complexity to generate them for all the layers is proportional to the number of neurons per layer and the number of classes, that is $O(LCn)$.

    The selection of frequencies ($f_c$) for each class is done to maintain sufficient distance among frequencies of different classes to avoid interference. 
    The range of available frequencies is defined by the length of $\mathbf{h}^{(l)}$. 
    Hence,frequencies can be assigned to be equally distributed in that range or randomly as long as they do not overlap. 
    In practice, we reduce the dimensions of $\mathbf{h}^{(l)}$ of convolutional layers by using average pooling before projecting it onto the basis $\mathbf{b}^{(l)}$. 
    This helps both in faster convergence of the method and in reducing the number of MAC operations.

    \subsection{Variations of LLS}\label{sec:variation_lls}
    So far, we have discussed LLS based on utilizing a basis of periodic vectors $\mathbf{b}^{(l)}$, generated from a fixed periodic function $g(\cdot)$. 
    However, such a base may not always be optimal for a given task.
    For instance, the amplitude of the vectors could be too large making it difficult for the algorithm to converge.
    Additionally, in problems with a large number of classes, the restriction to fixed periodic vectors may impede the model's ability to learn semantics in the data, such as grouping similar classes.
    
    To address these concerns, we propose two variations of LLS: LLS-M for learning the appropriate modulation of the fixed basis ($\mathbf{b}^{(l)}$), and LLS-MxM for learning to construct a new basis as a linear combination of the original fixed basis.
    
    \paragraph{LLS-M:} 
    In this variation, the new basis is simply a modulation of the original fixed basis, defined as $\mathbf{d}^{(l)} = \mathbf{M}^{(l)}\odot\mathbf{b}^{(l)}$, where $\mathbf{M}^{(l)}$ is a vector of trainable parameters with dimensions equal to the number of classes, i.e., $\mathbf{M}^{(l)}\in \mathbb{R}^C$. 
    Weight updates for LLS-M follow (\ref{eq:lls_update}), with $\mathbf{b}^{(l)}$ replaced by $\mathbf{d}^{(l)}$. 
    The updates for $\mathbf{M}^{(l)}$ are computed as follows:
    \begin{equation}
        \Delta \mathbf{M}^{(l)} = \frac{1}{N}\sum_{n=1}^N\mathbf{e}^{(l)}_n \odot(\mathbf{h}^{(l)}_n\mathbf{b}^{(l)\top})
        \label{eq:lls_m}
    \end{equation}
    
    \paragraph{LLS-MxM:}
    Here, the new basis vectors ($\mathbf{d}^{(l)}$) are obtained as a linear combination of the original fixed periodic vectors: 
    $\mathbf{d}^{(l)}=\mathbf{M}^{(l)}\mathbf{b}^{(l)}$ where $\mathbf{M}^{(l)}\in  \mathbb{R}^{C\times C}$.
    Weight updates are obtained following (\ref{eq:lls_update}), with the basis replaced by $\mathbf{d}^{(l)}$.
    Similar to LLS-M, updates for the matrix $\mathbf{M}^{(l)}$ are computed as follows:
    \begin{equation}
        \Delta \mathbf{M}^{(l)} = \frac{1}{N}\mathbf{e}^{(l)\top}_n(\mathbf{h}^{(l)}_n\mathbf{b}^{(l)\top})
        \label{eq:lls_mxm}
    \end{equation}

\section{Experimental evaluation}\label{sec:experimental_evaluation}
In this section, we assess the efficacy of LLS and its variations across several image classification datasets, which include MNIST \cite{LeCun2010MNISTDatabase}, FashionMNIST \cite{Xiao2017Fashion-mnist:Algorithms}, CIFAR10 \cite{Krizhevsky2009LearningImages}, CIFAR100 \cite{Krizhevsky2009LearningImages}, IMAGENETTE \cite{fast.ai2021Fastai/imagenette:French}, TinyIMAGENET \cite{Le2015TinyChallenge}, and Visual Wake Words (VWW) \cite{Chowdhery2019VisualDataset}.

We primarily evaluate the proposed learning rules using three models:  a 5-layer CNN (SmallConv), a VGG8 \cite{Nkland2019TrainingSignals}, and MobileNets-V1 (MBNet) \cite{Howard2017MobileNets:Applications}. 
Detailed descriptions of each model are provided in Appendix~\ref{appendix:models}. 
Additionally, information regarding hyperparameters, data pre-processing, and optimizer settings  is provided in Appendix~\ref{appendix:datasets}.

    \subsection{Effect of different basis in learning}\label{sec:different_basis}

    \begin{figure*}
    \centering
    \includegraphics[width=\textwidth]{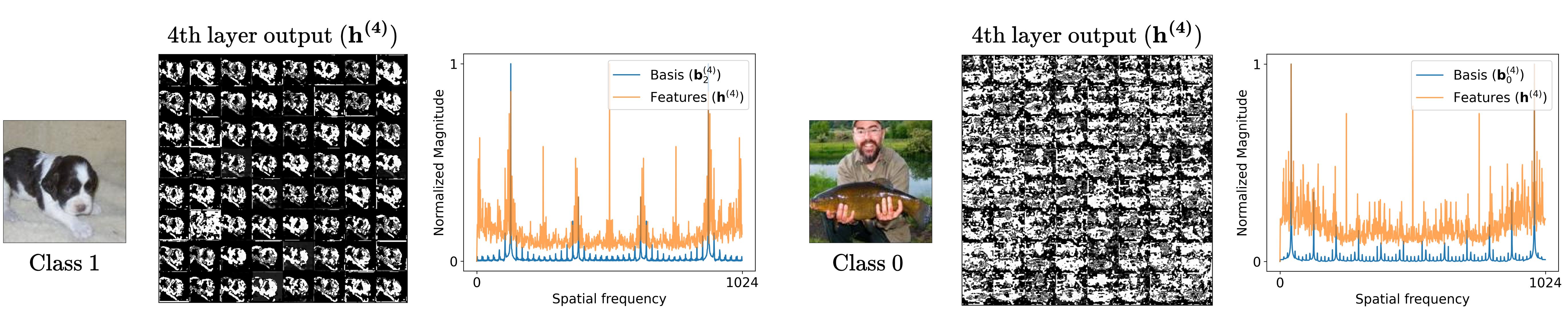}
    \caption{Neural activity synchronization induced by learning rule LLS\textsubscript{square} on the VGG8 model's 4th layer output ($\mathbf{h}^{(4)}$) for classes 0 and 1 from the IMAGENETTE dataset. The layer's response exhibits spatial periodicity coinciding with the periodic function selected as a basis ($\mathbf{b}^{(4)}$).}
    \label{fig:sync_act}
    \end{figure*}

    \begin{table*}[t]
        \renewcommand{\arraystretch}{0.7}
        \caption{LLS's performance comparison with different function $g(\cdot)$ to generate the basis $\mathbf{b}^{(l)}$. Test accuracy mean and std are reported over five trials. }
        \label{table:diff_waveform}
        \centering
        \resizebox{0.8\textwidth}{!}{
        \begin{tabular}{lccccc}
        \toprule
        \textbf{\begin{tabular}[c]{@{}c@{}}Function $g(.)$\end{tabular}} & \textbf{Model} &  \textbf{MNIST}  & \textbf{FashionMNIST} & \textbf{CIFAR10} & \textbf{IMAGENETTE} \\
        \midrule
        Cosine & \multirow{3}{*}{SmallConv} & $99.50\pm0.02$ & $89.57\pm0.22$ &$75.82\pm0.39$ & $78.03\pm0.35$ \\
        Square & & $99.50\pm0.02$ & $90.54\pm0.23$ & $77.79\pm0.31$ & $79.02\pm0.76$  \\
        Random & & $99.38\pm0.03$ & $87.30\pm0.18$ & $74.19\pm0.57$ & $71.70\pm1.45$  \\

        \midrule
        Cosine & \multirow{3}{*}{VGG8} & $99.52\pm0.02$ & $93.04\pm0.17$ & $86.92\pm0.27$ & $84.85\pm0.11$ \\
        Square &  & $99.54\pm0.01$ & $93.54\pm0.06$ & $88.64\pm0.12$ & $85.62\pm0.24$ \\
        Random & & $99.70\pm0.02$ & $93.77\pm0.08$ & $90.45\pm0.09$ & $87.09\pm0.28$ \\ 
        \bottomrule
        \end{tabular}
        }
    \end{table*}

    First, we compare the effect of different functions $g(\cdot)$ for generating the basis $\mathbf{b}^{(l)}$. 
    We consider two simple periodic functions: cosine ($g=\mathrm{cos}(f_ct)$) and square ($g=\mathrm{sign}(\mathrm{cos}(f_ct))$). 
    Both functions offer the advantage of being easily generated on-the-fly or require storage with minimal memory overhead due to their periodicity. 
    Additionally, we investigate the scenario where $g(\cdot)$ is a pseudo-random number generator, resulting in a random fixed vector $\mathbf{b}^{(l)}$.

    The results are evaluated on two models, SmallConv and VGG8, across four image classification datasets of increasing complexity.  
    Each model undergoes five training iterations with different random seeds, and the results are reported in Table~\ref{table:diff_waveform}.
    
    We observe that employing any of the three fixed vector bases with LLS yields high accuracy across all four vision tasks.
    Notably, for the SmallConv model, using LLS with a square basis function present the best accuracy results, followed by cosine basis. 
    In contrast, for the VGG8 model, the random basis exhibits better performance than the periodic basis, with square still performing better than cosine.
    This discrepancy may be attributed to the increased complexity of per-layer feature representations in deeper models, where a random vector offers more degrees of freedom for such representations. 
    However, it is important to note that a random vector is less hardware-friendly, as it requires specialized pseudo-random number generators, leading to energy and memory overhead, as discussed in \cite{Crafton2019DirectLearning}. 
    Therefore, in the subsequent sections, we primarily focus on LLS using a square $g(\cdot)$ function (LLS\textsubscript{square}).
    
    Moreover, employing a periodic function, such as a square function, induces layer neurons to synchronize with the frequency of the basis function. 
    This synchronization is demonstrated in Fig.~\ref{fig:sync_act}, where the activations for different classes align with the spatial frequencies of the basis function. 
    Here, the spectral decomposition is obtained by applying Fourier transform in the spatial dimension to both basis vectors ($\mathbf{b}^{(l)}$) and layer output activations ($\mathbf{h}^{(l)}$).
    As shown in Table~\ref{table:diff_waveform},  synchronization has a beneficial effect on accuracy for small models, such as SmallConv.
    A reason for this is that such models need to discriminate between classes by transforming inputs through only a few layers. 
    Thus, aligning the layers' outputs to periodic vectors might be easier than aligning random vectors.

    \subsection{Comparison with local learning algorithms}\label{sec:comparison_locallearning}
    In this section, we compare LLS\textsubscript{square} with other local learning methods that exhibit similar time and memory complexities.
    These methods include DFA \cite{Trondheim2016DirectNetworks}, DRTP \cite{Frenkel2021LearningNetworks}, and PEPITA \cite{Dellaferrera2022Error-drivenPass}. 
    For this comparison, we use the MNIST, CIFAR10 and CIFAR100 datasets, with results shown in Table~\ref{table:local_methods_comparison}.

    We observe that training the SmallConv model with DFA, DRTP or PEPITA resulted in low performance or did not converge at all. 
    For DFA, performance improved by increasing the number of channels threefold (SmallConvL).
    Consequently, we used SmallConvL for reporting results with BP and LLS.
    However, for DRTP and PEPITA, increasing number of channels did not yield satisfactory results, and hence, we opted for reporting accuracy of each task as reported in the original papers.
    
    As shown in Table~\ref{table:local_methods_comparison}, LLS demonstrates the best performance among the three local learning methods under consideration. 
    In terms of accuracy, LLS achieves results close to BP, while maintaining significantly lower time and memory complexities compared to BP.
    In fact, among all the methods in Table~\ref{table:local_methods_comparison}, only DRTP exhibit a time and memory complexities comparable to LLS. Furthermore, it is worth noting that while DFA, DRTP, and PEPITA do not scale well for deeper models and in many cases require wide DNNs to converge \cite{Song2021ConvergenceWeights}, LLS performs well on deeper models, as demonstrated in Section~\ref{sec:different_basis}.

    \begin{table*}[t]
        \renewcommand{\arraystretch}{0.7}
        \caption{Comparison with local learning algorithms (Test accuracy mean and std are reported) }
        \label{table:local_methods_comparison}
        \centering
        \resizebox{0.8\textwidth}{!}{
        \begin{tabular}{llllccc}
        \toprule
        \textbf{\begin{tabular}[c]{@{}c@{}}Method\end{tabular}}& \textbf{\begin{tabular}[c]{@{}c@{}}Time\end{tabular}}  &\textbf{\begin{tabular}[c]{@{}c@{}}Memory\end{tabular}}  & \textbf{Model} &  \textbf{MNIST}  & \textbf{CIFAR10} & \textbf{CIFAR100}  \\
        \midrule
        BP (baseline) & $O(Ln^2)$ & $O(Ln)$ & SmallConvL & $99.62\pm0.020$  &$87.57\pm0.13$ & $62.25\pm0.29$ \\
        \multirow{2}{*}{DFA} & \multirow{2}{*}{$O(LCn)$} & \multirow{2}{*}{$O(Ln)$} & SmallConvL & $97.90\pm0.17$  & $71.53\pm0.38$ & $44.93\pm0.52$  \\
        & & & \cite{Trondheim2016DirectNetworks} &$98.98\pm0.02$ & $73.10\pm0.50$ & $41.00\pm0.30$ \\
        DRTP & $O(LCn)$ & $O(n_{max})$ & \cite{Frenkel2021LearningNetworks} & $98.52\pm0.15$ & $68.96\pm0.45$ & $-$\\
        PEPITA & $O(Ln^2)$ & $O(n_{max})$ & \cite{Dellaferrera2022Error-drivenPass} & $98.29\pm0.13$  & $56.33\pm1.35$ & $27.56\pm0.60$ \\ 
        LLS\textsubscript{square} (Ours) & $O(LCn)$ & $O(n_{max})$ & SmallConvL & $99.57\pm0.03$ &  $84.10\pm0.27$ &  $55.32\pm0.38$\\
        \bottomrule
        \end{tabular}
        }
    \end{table*}

    \begin{table*}[t]
        \renewcommand{\arraystretch}{0.6}
        \caption{Performance comparison on image classification datasets. Accuracy mean and std are reported over five trials, the additional params refers to additional trainable parameters, and  \#MAC is estimated for the number of ops required to generate the learning signal ($\frac{\partial\mathcal{L}}{\partial \mathbf{h}^{(l)}}$). }
        \label{table:performace_comparison}
        \centering
        \resizebox{0.7\textwidth}{!}{
        \begin{tabular}{lccrrr}
        \toprule
        \multicolumn{1}{c}{\textbf{Method}} & \multicolumn{1}{c}{\textbf{Model}} &  \textbf{\begin{tabular}[c]{@{}c@{}}Accuracy \\ (mean$\pm$std)\end{tabular}} & \textbf{\begin{tabular}[c]{@{}c@{}}\# MAC$^1$\\ ($\times 10^6$)\end{tabular}} & \textbf{\begin{tabular}[c]{@{}c@{}}Memory$^1$\\ (MB)\end{tabular}} & \textbf{\begin{tabular}[c]{@{}c@{}}Additional \\params\end{tabular}}\\ 
        \midrule        
        \multicolumn{5}{c}{CIFAR10} \\ \midrule
        \multicolumn{1}{l}{BP}  & \multicolumn{1}{c}{VGG8}& $94.12\pm0.12$ & $719.33$ & $1082$ & -\\ 
        \multicolumn{1}{l}{Local Losses} & \multicolumn{1}{c}{VGG8}  & $91.93\pm0.07$ & $2.56$ & $576$ & $1.02\times10^5$\\ 
        \multicolumn{1}{l}{LLS\textsubscript{square} (Ours)} & \multicolumn{1}{c}{VGG8} & $88.64\pm0.12$ & $2.46$ & $574$ & $0$\\ 
        \multicolumn{1}{l}{LLS-M\textsubscript{square} (Ours)} & \multicolumn{1}{c}{VGG8} & $90.43\pm0.24$ & $2.46$ & $574$ & $70$\\ 
        \multicolumn{1}{l}{LLS-MxM\textsubscript{square} (Ours)} & \multicolumn{1}{c}{VGG8} & $90.89\pm0.09$ & $2.46$ & $574$ & $700$\\ 
        \midrule
        \multicolumn{5}{c}{IMAGENETTE} \\ \midrule
        \multicolumn{1}{l}{BP}  & \multicolumn{1}{c}{VGG8}& $90.92\pm0.27$ & $11477.81$ & $15858$ & -\\ 
        \multicolumn{1}{l}{Local Losses} & \multicolumn{1}{c}{VGG8}  & $88.06\pm0.12$ & $36.48$ & $7319$ & $1.02\times10^5$\\ 
        \multicolumn{1}{l}{LLS\textsubscript{square} (Ours)} & \multicolumn{1}{c}{VGG8} & $85.62\pm0.24$ & $36.38$ & $7318$ & $0$\\ 
        \multicolumn{1}{l}{LLS-M\textsubscript{square} (Ours)} & \multicolumn{1}{c}{VGG8} & $86.60\pm0.37$ & $36.38$ & $7318$ & $70$\\ 
        \multicolumn{1}{l}{LLS-MxM\textsubscript{square} (Ours)} & \multicolumn{1}{c}{VGG8} & $87.29\pm0.29$ & $36.38$ & $7319$ & $700$\\ 
        \midrule
        \multicolumn{5}{c}{CIFAR100} \\ \midrule
        \multicolumn{1}{l}{BP}  & \multicolumn{1}{c}{VGG8}& $73.69\pm0.39$ & $719.40$ & $1083$ & -\\ 
        \multicolumn{1}{l}{Local Losses} & \multicolumn{1}{c}{VGG8}  & $69.26\pm0.36$ & $5.33$ & $598$ & $1.02\times 10^6$\\ 
        \multicolumn{1}{l}{LLS\textsubscript{square} (Ours)} & \multicolumn{1}{c}{VGG8} & $58.84\pm0.33$ & $4.30$ & $577$ & $0$ \\ 
        \multicolumn{1}{l}{LLS-M\textsubscript{square} (Ours)} & \multicolumn{1}{c}{VGG8} & $62.55\pm0.24$ & $4.31$ & $577$ & $700$ \\ 
        \multicolumn{1}{l}{LLS-MxM\textsubscript{square} (Ours)} & \multicolumn{1}{c}{VGG8} & $68.81\pm0.19$ & $4.51$ & $578$ & $0.70\times 10^5$ \\ 
        \midrule
        
        \multicolumn{5}{c}{TinyIMAGENET} \\ \midrule
        \multicolumn{1}{l}{BP}  & \multicolumn{1}{c}{VGG8}& $61.10\pm0.25$ & $2871.20$ & $4048$ & - \\ 
        \multicolumn{1}{l}{Local Losses} & \multicolumn{1}{c}{VGG8}  & $54.00\pm0.11$ & $15.18$ & $1971$ & $2.04\times 10^6$\\ 
        \multicolumn{1}{l}{LLS\textsubscript{square} (Ours)} & \multicolumn{1}{c}{VGG8} & $35.99\pm0.38$ & $13.13$ & $1928$ & $0$ \\ 
        \multicolumn{1}{l}{LLS-M\textsubscript{square} (Ours)} & \multicolumn{1}{c}{VGG8} & $41.89\pm0.20$ & $13.14$ & $1928$ & $1400$\\ 
        \multicolumn{1}{l}{LLS-MxM\textsubscript{square} (Ours)} & \multicolumn{1}{c}{VGG8} & $51.41\pm0.48$ & $13.97$ & $1932$ & $0.28\times10^6$\\ 
        \midrule
        \multicolumn{5}{c}{Visual Wake Words (VWW)} \\ \midrule
        \multicolumn{1}{l}{BP} & \multicolumn{1}{c}{MBNet} & $88.49\pm0.28$ & $181.83$ & $3036$ & -\\
        \multicolumn{1}{l}{Local Losses} & \multicolumn{1}{c}{MBNet} & $82.49\pm0.17$ & $178.28$ & $730$ & $0.28\times10^5$\\
        \multicolumn{1}{l}{LLS\textsubscript{square} (Ours)} & \multicolumn{1}{c}{MBNet} & $81.91\pm0.16$ & $178.23$ & $729$ & $0$\\
        \multicolumn{1}{l}{LLS-M\textsubscript{square} (Ours)} & \multicolumn{1}{c}{MBNet} & $82.71\pm0.42$ & $178.23$ & $729$ & $28$\\
        \multicolumn{1}{l}{LLS-MxM\textsubscript{square} (Ours)} & \multicolumn{1}{c}{MBNet} & $83.66\pm0.21$ & $178.25$ & $729$ & $560$\\
        \midrule
        \multicolumn{6}{l}{\small{$^1$: \# MAC is estimated for a batch size of 1 and GPU memory is measured for a batch size of 128.}}\\
        \end{tabular}
        }
    \end{table*}

    \subsection{Performance comparison on deeper models}\label{sec:performance_comparison}

    In this section, we conduct a performance comparison of LLS and its variations on five image classification datasets: CIFAR10, CIFAR100, IMAGENETTE, TinyIMAGENET, and VWW. 
    These datasets cover a wide range of classification tasks, including low to high-resolution images and tasks with few to multiple classes. 
    Notably, we emphasize the experiments conducted on the VWW dataset, as it holds significance for edge vision applications and serves as a relevant use case for on-device learning \cite{Chowdhery2019VisualDataset}.
    The comparison considers four metrics: accuracy, the number of MAC operations required to compute the learning signal, the peak memory usage, and the number of additional trainable parameters needed by each method. 
    We compare our method against BP and the local losses method \cite{Nkland2019TrainingSignals}. 
    Note,  local losses method employs a linear classifier per layer.
    
    \paragraph{CIFAR10 and IMAGENETTE}
    First, we examine tasks with a few number of classes and different image resolutions, such as CIFAR10 and IMAGENETTE. 
    As depicted in Table~\ref{table:performace_comparison}, LLS achieves high accuracy, closely following BP and Local Losses.    
    Note, that LLS achieves such high accuracy with approximately $300\times$ fewer MAC operations and half the memory usage compared to BP, and without requiring additional trainable parameters. 
    To further narrow the accuracy gap, we explore variations of LLS, such as LLS-M and LLS-MxM.
    Both variations improve the accuracy to be closer to BP with almost no increase in MACs and memory usage. 
    Note, however, the accuracy improvement comes at the cost of employing some additional trainable parameters. 
    It is important to note that LLS-MxM still requires approximately $100\times$ fewer trainable parameters than Local Losses.

    \begin{figure*}[h]
    \centering
    \includegraphics[width=\textwidth]{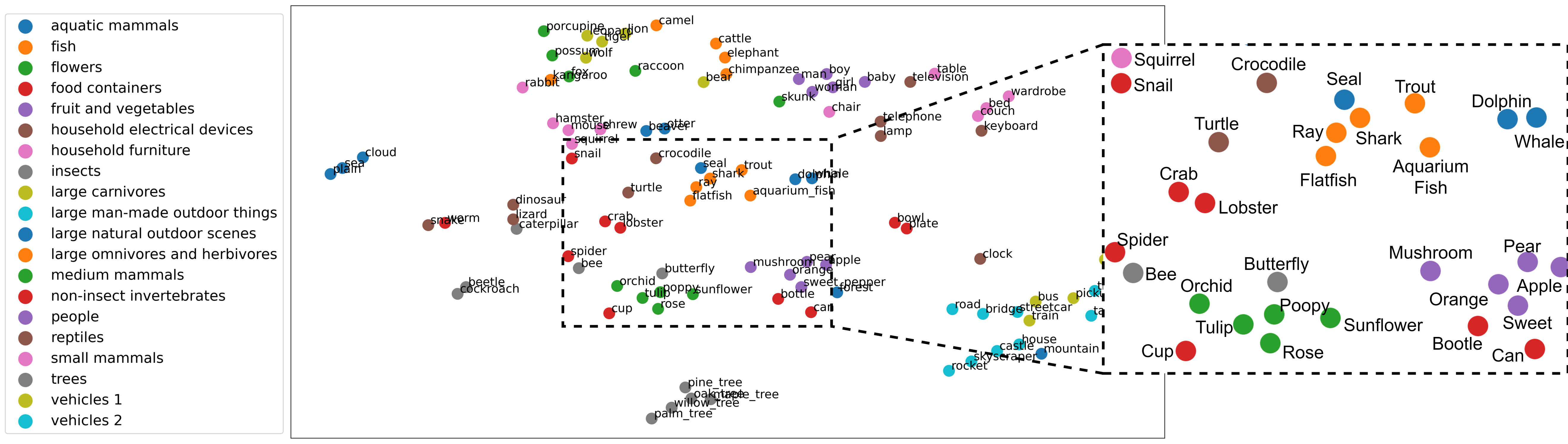}
    \caption{Projection of the linear combination matrix $\mathbf{M}^{(l)}$ of the fixed basis $\mathbf{b}^{(l)}$ using t-SNE. $\mathbf{M}^{(l)}$ is obtained after training a VGG8 model with LLS-MxM on CIFAR100. The results provide evidence that our learning rule can learn better basis (as a linear combination of a fixed basis) and can encode semantics within it. Points are colored using the twenty super-class labels provided in CIFAR100.}
    \label{fig:semantics_cifar100}
    \end{figure*}

    \begin{figure*}[t]
    \centering
    \includegraphics[width=\textwidth]{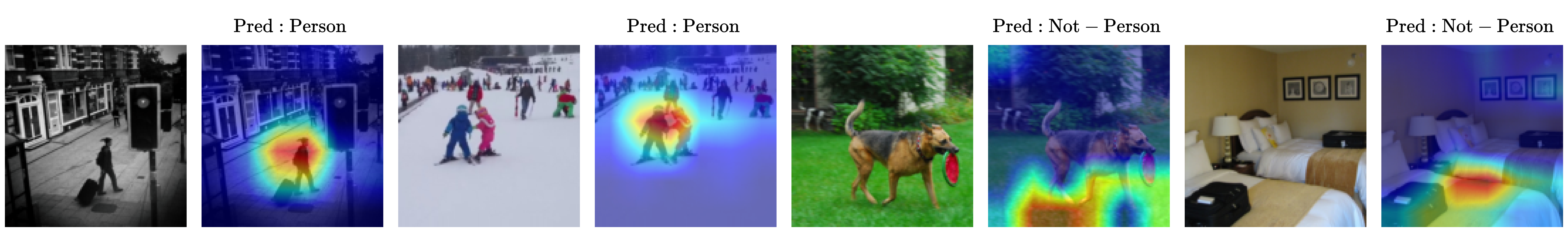}
    \caption{Visual explanations, obtained with the Grad-CAM method, for predictions of the MBNet model trained with LLS-MxM on the VWW dataset. It can be observed that our method allows the model to learn high level image features to discern about the presence of a person or not in an image.}
    \label{fig:vww_gradcam}
    \end{figure*}

    \paragraph{CIFAR100 and TinyIMAGENET}\label{sec:exp_cifar100}
    For tasks with hundreds of classes such as CIFAR100 and TinyIMAGENET, LLS exhibits significant accuracy drop compared to BP. 
    This is attributed to the orthogonal nature of the periodic vectors, which compels the model to represent each class orthogonally, even when semantically some classes have similar representations. 
    Essentially, the basic form of LLS may not effectively capture semantics. 
    Additionally, increasing the number of classes also increases the number of frequencies used to generate the fixed basis, leading to overlapping frequencies.
    We applied LLS-M learning for the above problems. 
    LLS-M improves the accuracy, but only marginally, as the problems associated with orthogonality of the bases could not be completely solved by simply modulating the bases. 
    In contrast, LLS-MxM learns to create a better basis as a linear combination of the original basis, offering a larger improvement and bringing the accuracy closer to BP, as show in Table~\ref{table:performace_comparison}. 
    To further verify that LLS-MxM can actually learn semantics, we analyze the learned linear combination matrix ($\mathbf{M}^{(l)}$) used to create the new basis.
    For instance, for a VGG8 model trained on CIFAR100, we project the $\mathbf{M}^{(l)}$ matrix into a 2D space using t-SNE \cite{Maaten2008VisualizingT-SNE} using the twenty super-classes provided in the dataset as ground truth. 
    The results of this projection are illustrated in Fig.~\ref{fig:semantics_cifar100}, wherein vectors representing similar classes are grouped together. 
    The accuracy improvements shown in Table~\ref{table:performace_comparison} and the clustering of similar classes illustrated in Fig.~\ref{fig:semantics_cifar100} demonstrate the ability of LLS-MxM to encode semantic knowledge in the formation of the new basis.
    Furthermore, it is worth noting that LLS-MxM requires approximately $200\times$ fewer MACs and half memory compared to BP, and approximately $10\times$ fewer trainable parameters than Local Losses.

    \paragraph{Visual Wake Words (VWW)}
    
    Since our learning rule targets on-device learning scenarios, we tested the method on the VWW dataset using a MobileNetsV1 model. 
    Note, the task and the model are suitable for on-device learning. 
    The results are shown in Table~\ref{table:performace_comparison}. 
    For this task, LLS-M and LLS-MxM outperforms the Local Losses method in all metrics (accuracy, MACs, memory, and trainable parameters). 
    Compared to BP, LLS, LLS-M and LLS-MxM show competitive accuracy with fewer MACs and $4\times$ lower memory usage.
    Moreover, to understand the model's learning ability, we used the Grad-CAM method \cite{Selvaraju2017Grad-CAM:Localization} to obtain visual explanations of the parts of the image most relevant for a particular prediction. 
    As shown in Fig.~\ref{fig:vww_gradcam}, the MBNet model trained with LLS-MxM successfully learns high-level image features indicative of the presence of people in a given frame. 
    This provides evidence that our method allows the model to learn complex representations.

\section{Conclusions}\label{sec:conclusions}
In this work, we introduced a novel local learning rule, LLS, inspired by the synchronization of neural activity observed in biological systems, which is associated with memory formation and cognitive learning. 
LLS utilizes fixed periodic basis vectors to synchronize the activity of neurons within the same layer. 
Moreover, the deliberate choice of simple periodic functions, such as cosine and square functions, enables the generation of such basis easily and on-the-fly on low-power devices without imposing significant hardware overhead.
Experimental validation demonstrates that LLS and its variations (LLS-M and LLS-MxM) achieve high accuracy comparable to BP across various image classification datasets, including CIFAR10, CIFAR100, IMAGENETTE, TinyIMAGENET, and VWW. 
Remarkably, this high accuracy is attained with significantly fewer MAC operations, reduced memory usage, and a minimal number of additional trainable parameters.
Furthermore, employing the Grad-CAM method for visual explanations reveals that LLS and its variants can capture high-level information relevant to predictions. 
In summary, the demonstrated high accuracy and efficiency of LLS make it well-suited for on-device learning applications, particularly in scenarios where computational resources are severely constrained.

\section*{Acknowledgments}
This work was supported in part by the Center for Co-design of Cognitive Systems (CoCoSys), one of the seven centers in JUMP 2.0, a Semiconductor Research Corporation (SRC) program, and in part by the Department of Energy (DoE).

{\small
\bibliographystyle{ieee_fullname}
\bibliography{references}

\begin{thebibliography}{10}\itemsep=-1pt

\bibitem{Ankit2020PANTHER:ReRAM}
Aayush Ankit, Izzat~El Hajj, Sai~Rahul Chalamalasetti, Sapan Agarwal, Matthew Marinella, Martin Foltin, John~Paul Strachan, Dejan Milojicic, Wen~Mei Hwu, and Kaushik Roy.
\newblock {PANTHER: A Programmable Architecture for Neural Network Training Harnessing Energy-Efficient ReRAM}.
\newblock {\em IEEE Transactions on Computers}, 69(8):1128--1142, 8 2020.

\bibitem{Belilovsky2018GreedyImageNet}
Eugene Belilovsky, Michael Eickenberg, and Edouard Oyallon.
\newblock {Greedy Layerwise Learning Can Scale to ImageNet}.
\newblock In {\em International Conference on Machine Learning}, 2018.

\bibitem{Carrillo-Reid2019ControllingEnsembles}
Luis Carrillo-Reid, Shuting Han, Weijian Yang, Alejandro Akrouh, and Rafael Yuste.
\newblock {Controlling Visually Guided Behavior by Holographic Recalling of Cortical Ensembles}.
\newblock {\em Cell}, 178(2):447--457, 7 2019.

\bibitem{Chowdhery2019VisualDataset}
Aakanksha Chowdhery, Pete Warden, Jonathon Shlens, Andrew Howard, and Rocky Rhodes.
\newblock {Visual Wake Words Dataset}.
\newblock {\em arXiv: 1906.05721}, 6 2019.

\bibitem{Crafton2019DirectLearning}
Brian Crafton, Abhinav Parihar, Evan Gebhardt, and Arijit Raychowdhury.
\newblock {Direct feedback alignment with sparse connections for local learning}.
\newblock {\em Frontiers in Neuroscience}, 13(MAY), 2019.

\bibitem{Defazio2024Schedule-FreeTrain}
Aaron Defazio, Xingyu Yang, Konstantin Mishchenko, Ashok Cutkosky, Harsh Mehta, and Ahmed Khaled.
\newblock {Schedule-Free Learning - A New Way to Train}.
\newblock https://github.com/facebookresearch/schedule{\_}free, 2024.

\bibitem{Dellaferrera2022Error-drivenPass}
Giorgia Dellaferrera, Gabriel Kreiman, and Gabriel Kreiman.
\newblock {Error-driven Input Modulation: Solving the Credit Assignment Problem without a Backward Pass}.
\newblock In Kamalika Chaudhuri, Stefanie Jegelka, Le Song, Csaba Szepesvari, Gang Niu, and Sivan Sabato, editors, {\em Proceedings of the 39th International Conference on Machine Learning}, pages 4937--4955. PMLR, 7 2022.

\bibitem{fast.ai2021Fastai/imagenette:French}
{fast.ai}.
\newblock {fastai/imagenette: A smaller subset of 10 easily classified classes from Imagenet, and a little more French}, 2021.

\bibitem{Frenkel2021LearningNetworks}
Charlotte Frenkel, Martin Lefebvre, and David Bol.
\newblock {Learning Without Feedback: Fixed Random Learning Signals Allow for Feedforward Training of Deep Neural Networks}.
\newblock {\em Frontiers in Neuroscience}, 15:629892, 2 2021.

\bibitem{GuevaraErra2017NeuralDynamics}
Ramon Guevara~Erra, Jose~L Perez~Velazquez, and Michael Rosenblum.
\newblock {Neural synchronization from the perspective of non-linear dynamics}.
\newblock {\em Frontiers in computational neuroscience}, 11:98, 2017.

\bibitem{Hinton2022TheInvestigations}
Geoffrey Hinton.
\newblock {The Forward-Forward Algorithm: Some Preliminary Investigations}.
\newblock Technical report, 2022.

\bibitem{Hopfield1982NeuralAbilities.}
J.~J. Hopfield.
\newblock {Neural networks and physical systems with emergent collective computational abilities.}
\newblock {\em Proceedings of the National Academy of Sciences of the United States of America}, 79(8):2554, 1982.

\bibitem{Howard2017MobileNets:Applications}
Andrew~G. Howard, Menglong Zhu, Bo Chen, Dmitry Kalenichenko, Weijun Wang, Tobias Weyand, Marco Andreetto, and Hartwig Adam.
\newblock {MobileNets: Efficient Convolutional Neural Networks for Mobile Vision Applications}.
\newblock {\em arXiv preprint arXiv:1704.04861}, 4 2017.

\bibitem{Journe2023HebbianFeedback}
Adrien Journ{\'{e}}, Hector Garcia~Rodriguez, Qinghai Guo, and Timoleon Moraitis.
\newblock {Hebbian Deep Learning Without Feedback}.
\newblock In {\em 2023 International Conference on Learning Representations}, 2023.

\bibitem{Jutras2010SynchronousFormation}
Michael~J Jutras and Elizabeth~A Buffalo.
\newblock {Synchronous neural activity and memory formation}.
\newblock {\em Current opinion in neurobiology}, 20(2):150--155, 2010.

\bibitem{Krizhevsky2009LearningImages}
Alex Krizhevsky.
\newblock {Learning Multiple Layers of Features from Tiny Images}.
\newblock 2009.

\bibitem{Le2015TinyChallenge}
Ya Le and Xuan~S Yang.
\newblock {Tiny ImageNet Visual Recognition Challenge}.
\newblock 2015.

\bibitem{LeCun2010MNISTDatabase}
Yann LeCun, Corinna Cortes, and C~J Burges.
\newblock {MNIST handwritten digit database}.
\newblock {\em ATT Labs [Online]. Available: http://yann.lecun.com/exdb/mnist}, 2, 2010.

\bibitem{Lillicrap2016RandomLearning}
Timothy~P. Lillicrap, Daniel Cownden, Douglas~B. Tweed, and Colin~J. Akerman.
\newblock {Random synaptic feedback weights support error backpropagation for deep learning}.
\newblock {\em Nature Communications}, 7, 11 2016.

\bibitem{Lillicrap2020BackpropagationBrain}
Timothy~P. Lillicrap, Adam Santoro, Luke Marris, Colin~J. Akerman, and Geoffrey Hinton.
\newblock {Backpropagation and the brain}.
\newblock {\em Nature Reviews Neuroscience}, 21(6):335--346, 6 2020.

\bibitem{Maaten2008VisualizingT-SNE}
Laurens van~der Maaten and Geoffrey Hinton.
\newblock {Visualizing Data using t-SNE}.
\newblock {\em Journal of Machine Learning Research}, 9(86):2579--2605, 2008.

\bibitem{Miller2014VisualEnsembles}
Jae Eun~Kang Miller, Inbal Ayzenshtat, Luis Carrillo-Reid, and Rafael Yuste.
\newblock {Visual stimuli recruit intrinsically generated cortical ensembles}.
\newblock {\em Proceedings of the National Academy of Sciences of the United States of America}, 111(38):E4053--E4061, 9 2014.

\bibitem{Nkland2019TrainingSignals}
Arild N{\o}kland and Lars~H Eidnes.
\newblock {Training Neural Networks with Local Error Signals}.
\newblock In {\em Proceedings of the 36 th International Conference on Machine Learning}, 2019.

\bibitem{Ororbia2023Backpropagation-FreeAlignment}
Alexander~G. Ororbia, Ankur Mali, Daniel Kifer, and C. Lee~Giles.
\newblock {Backpropagation-Free Deep Learning with Recursive Local Representation Alignment}.
\newblock {\em Proceedings of the AAAI Conference on Artificial Intelligence}, 37(8):9327--9335, 6 2023.

\bibitem{Peng2021NeuroSimV2}
Xiaochen Peng, Shanshi Huang, Hongwu Jiang, Anni Lu, and Shimeng Yu.
\newblock {DNN+NeuroSim V2.0: An End-to-End Benchmarking Framework for Compute-in-Memory Accelerators for On-Chip Training}.
\newblock {\em IEEE Transactions on Computer-Aided Design of Integrated Circuits and Systems}, 40(11):2306--2319, 11 2021.

\bibitem{Selvaraju2017Grad-CAM:Localization}
Ramprasaath~R. Selvaraju, Michael Cogswell, Abhishek Das, Ramakrishna Vedantam, Devi Parikh, and Dhruv Batra.
\newblock {Grad-CAM: Visual Explanations from Deep Networks via Gradient-Based Localization}.
\newblock {\em Proceedings of the IEEE International Conference on Computer Vision}, 2017-October:618--626, 12 2017.

\bibitem{Song2021ConvergenceWeights}
Ganlin Song, Ruitu Xu, and John Lafferty.
\newblock {Convergence and Alignment of Gradient Descent with Random Backpropagation Weights}.
\newblock In {\em 35th Conference on Neural Information Processing Systems (NeurIPS 2021)}, 2021.

\bibitem{Trondheim2016DirectNetworks}
Arild~Nøkland Trondheim.
\newblock {Direct Feedback Alignment Provides Learning in Deep Neural Networks}.
\newblock {\em Advances in Neural Information Processing Systems}, 29, 2016.

\bibitem{Wang2021RevisitingTraining}
Yulin Wang, Zanlin Ni, Shiji Song, Le Yang, and Gao Huang.
\newblock {Revisiting Locally Supervised Learning: an Alternative to End-to-end Training}.
\newblock In {\em International Conference on Learning Representations}, 2021.

\bibitem{Xiao2017Fashion-mnist:Algorithms}
Han Xiao, Kashif Rasul, and Roland Vollgraf.
\newblock {Fashion-mnist: a novel image dataset for benchmarking machine learning algorithms}.
\newblock {\em arXiv preprint arXiv:1708.07747}, 2017.

\bibitem{Zhang2018SignSystems}
Qingtian Zhang, Huaqiang Wu, Peng Yao, Wenqiang Zhang, Bin Gao, Ning Deng, and He Qian.
\newblock {Sign backpropagation: An on-chip learning algorithm for analog RRAM neuromorphic computing systems}.
\newblock {\em Neural Networks}, 108:217--223, 12 2018.

\bibitem{Zhong2017RandomAugmentation}
Zhun Zhong, Liang Zheng, Guoliang Kang, Shaozi Li, and Yi Yang.
\newblock {Random Erasing Data Augmentation}.
\newblock {\em AAAI 2020 - 34th AAAI Conference on Artificial Intelligence}, pages 13001--13008, 8 2017.

\end{thebibliography}
}

\appendix

\section{Experimental Setup}\label{appendix:setup}
In this section, we describe the architecture of all models used in this work, the datasets and preprocessing operations, the training details including hyperparameters for each experiment, and the compute resources employed.
    
    \subsection{Model architecture}\label{appendix:models}
    In this work, we use four models: SmallConv, SmallConvL, VGG8 \cite{Nkland2019TrainingSignals}, and MobileNetV1 \cite{Howard2017MobileNets:Applications}. 
    These models are built using the following three basic blocks: ConvBlock, ConvDWBlock, and LinearBlock.
    \begin{itemize}
        \item ConvBlock is composed of three layers in the following order: a convolutional layer (Conv), a batch normalization layer (BN), and a Leaky ReLU (LeakyReLU).
        \item ConvDWBlock is composed of five layers in the following order: a depthwise convolutional layer (ConvDW), a BN layer, a Conv layer with kernel size of 1 (Conv1x1), another BN layer, and a LeakyReLU layer.
        \item LinearBlock is composed of three layers: a fully-connected layer (Linear), a BN layer, and a LeakyReLU.
    \end{itemize}
    The architecture of each of the models is described in Table~\ref{table:model_arch}.
    Note that LLS was applied at the outputs of each ConvBlock, ConvDWBlock, and LinearBlock, after the output dimensions were reduced to a size of 2048 (or lower depending on the output dimensions) using an Adaptive Average Pooling (AdaptiveAvgPool) layer.

    \begin{table*}[]
    \renewcommand{\arraystretch}{0.9}
    \caption{Model architectures. For the ConvBlock and ConvDWBlock A,B,C means A means the kernel size, B the number of output channels and C the stride. For Linear Block, A means the number of output neurons.}
    \label{table:model_arch}
    \centering
    \resizebox{0.65\textwidth}{!}{%
    \begin{tabular}{cllll}
    \toprule
    \multicolumn{1}{c}{\textbf{ID}} & \multicolumn{1}{c}{\textbf{SmallConv}} & \multicolumn{1}{c}{\textbf{SmallConvL}} & \multicolumn{1}{c}{\textbf{VGG8}} & \multicolumn{1}{c}{\textbf{MobileNetV1}} \\ 
    \midrule
    1 & \begin{tabular}[c]{@{}l@{}}ConvBlock \\ 3, 32, 1\end{tabular} & \begin{tabular}[c]{@{}l@{}}ConvBlock \\ 3, 96, 1\end{tabular} & \begin{tabular}[c]{@{}l@{}}ConvBlock\\ 3, 128, 1\end{tabular} & \begin{tabular}[c]{@{}l@{}}ConvBlock \\ 3, 32, 2\end{tabular} \\
    \midrule
    2 & \begin{tabular}[c]{@{}l@{}}MaxPool \\ 2, 2\end{tabular} & \begin{tabular}[c]{@{}l@{}}MaxPool \\ 2, 2\end{tabular} & \begin{tabular}[c]{@{}l@{}}ConvBlock\\ 3, 256, 1\end{tabular} & \begin{tabular}[c]{@{}l@{}}ConvDWBlock \\ 3, 64, 1\end{tabular} \\
    \midrule
    3 & \begin{tabular}[c]{@{}l@{}}ConvBlock \\ 3, 64, 1\end{tabular} & \begin{tabular}[c]{@{}l@{}}ConvBlock\\ 3, 192, 1\end{tabular} & \begin{tabular}[c]{@{}l@{}}MaxPool\\ 2, 2\end{tabular} & \begin{tabular}[c]{@{}l@{}}ConvDWBlock \\ 3, 128, 2\end{tabular} \\ 
    \midrule
    4 & \begin{tabular}[c]{@{}l@{}}MaxPool \\ 2, 2\end{tabular} & \begin{tabular}[c]{@{}l@{}}MaxPool \\ 2, 2\end{tabular} & \begin{tabular}[c]{@{}l@{}}ConvBlock\\ 3, 256, 1\end{tabular} & \begin{tabular}[c]{@{}l@{}}ConvDWBlock \\ 3, 128, 1\end{tabular} \\
    \midrule
    5 & \begin{tabular}[c]{@{}l@{}}ConvBlock \\ 3, 128, 1\end{tabular} & \begin{tabular}[c]{@{}l@{}}ConvBlock\\ 3, 512, 1\end{tabular} & \begin{tabular}[c]{@{}l@{}}ConvBlock\\ 3, 256, 1\end{tabular} & \begin{tabular}[c]{@{}l@{}}ConvDWBlock \\ 3, 256, 2\end{tabular} \\
    \midrule
    6 & \begin{tabular}[c]{@{}l@{}}AdaptiveAvgPool \\ (2, 2)\end{tabular} & \begin{tabular}[c]{@{}l@{}}AdaptiveAvgPool \\ (2, 2)\end{tabular} & \begin{tabular}[c]{@{}l@{}}Max Pool \\ 2, 2\end{tabular} & \begin{tabular}[c]{@{}l@{}}ConvDWBlock \\ 3, 256, 1\end{tabular} \\
    \midrule
    7 & \begin{tabular}[c]{@{}l@{}}LinearBlock \\ 512\end{tabular} & \begin{tabular}[c]{@{}l@{}}LinearBlock \\ 1024\end{tabular} & \begin{tabular}[c]{@{}l@{}}ConvBlock \\ 3, 512, 1\end{tabular} & \begin{tabular}[c]{@{}l@{}}ConvDWBlock \\ 3, 512, 2\end{tabular} \\
    \midrule
    8 & - & - & \begin{tabular}[c]{@{}l@{}}ConvBlock\\ 3, 512, 1\end{tabular} & \begin{tabular}[c]{@{}l@{}}ConvDWBlock \\ 3, 512, 1\end{tabular} \\
    \midrule
    9 & - & - & \begin{tabular}[c]{@{}l@{}}AdaptiveAvgPool \\ (2, 2)\end{tabular} & \begin{tabular}[c]{@{}l@{}}ConvDWBlock \\ 3, 512, 1\end{tabular} \\
    \midrule
    10 & - & - & \begin{tabular}[c]{@{}l@{}}LinearBlock \\ 1024\end{tabular} & \begin{tabular}[c]{@{}l@{}}ConvDWBlock \\ 3, 512, 1\end{tabular} \\
    \midrule
    11 & - & - & - & \begin{tabular}[c]{@{}l@{}}ConvDWBlock \\ 3, 512, 1\end{tabular} \\
    \midrule
    12 & - & - & - & \begin{tabular}[c]{@{}l@{}}ConvDWBlock \\ 3, 512, 1\end{tabular} \\
    \midrule
    13 & - & - & - & \begin{tabular}[c]{@{}l@{}}ConvDWBlock \\ 3, 1024, 2\end{tabular} \\
    \midrule
    14 & - & - & - & \begin{tabular}[c]{@{}l@{}}ConvDWBlock \\ 3, 1024, 1\end{tabular} \\
    \midrule
    15 & - & - & - & \begin{tabular}[c]{@{}l@{}}AdaptiveAvgPool \\ (2, 2)\end{tabular} \\ 
    \bottomrule
    \end{tabular}%
    }
    \end{table*}

   \subsection{Datasets}\label{appendix:datasets}
    In this section, we provide a brief description of the datasets used in this work: MNIST \cite{LeCun2010MNISTDatabase}, FashionMNIST \cite{Xiao2017Fashion-mnist:Algorithms}, CIFAR10 \cite{Krizhevsky2009LearningImages}, CIFAR100 \cite{Krizhevsky2009LearningImages}, IMAGENETTE \cite{fast.ai2021Fastai/imagenette:French}, TinyIMAGENET \cite{Le2015TinyChallenge}, and Visual Wake Words (VWW) \cite{Chowdhery2019VisualDataset}.
    
    \paragraph{MNIST:} This dataset consists of 70000 grayscale images of handwritten digits (0-9), each of size 28x28 pixels. It is divided into 60000 training images and 10,000 test images.
    
    \paragraph{FashionMNIST:} This dataset consists of 70000 grayscale images of fashion items, such a clothing and accessories, each of size 28x28 pixels. Similar to MNIST, it is divided into 60,000 training images and 10000 test images.
    
    \paragraph{CIFAR10:} This dataset consists of 60000 color images in 10 different classes, with each class containing 6000 images. The images are 32x32 pixels in size and the dataset is split into 50000 training images and 10000 test images.
    
    \paragraph{CIFAR100:}  It is similar to CIFAR-10 but contains 100 classes with 600 images per class. The images are each of size 32x32 pixels. The dataset is divided into 50000 training images and 10,000 test images. Each class has 500 training images and 100 test images. Additionally, CIFAR-100 includes labels for twenty super-classes, each grouping together five similar classes, providing a hierarchical structure for more detailed analysis.
    
    \paragraph{IMAGENETTE}
    This dataset is a subset of the larger ImageNet dataset, containing 10 easily classified classes such as tench, English springer, cassette player, chain saw, church, French horn, garbage truck, gas pump, golf ball, and parachute. It consists of 13000 images each with a resolution of 160x160 pixels.
    
    \paragraph{TinyIMAGENET}
    This dataset is a scaled-down version of the ImageNet dataset, containing 200 classes with 500 training images, 50 validation images, and 50 test images per class. The images are resized to 64x64 pixels.
    
    \paragraph{Visual Wake Words (VWW):} This dataset is designed for tiny, low-power computer vision models. It contains images labeled with the presence or absence of a person. The images are resized to 128x128 pixels. The dataset is divided into 115000 training images and 8000 test images.
    
    These datasets provide a diverse range of image classification challenges, facilitating the evaluation of models across various levels of complexity and application scenarios.

    \subsection{Training Details}
    All models reported in this work were trained with a batch size of 128 using the Schedule-Free AdamW optimizer \cite{Defazio2024Schedule-FreeTrain} with a learning rate of $5\times 10^{-3}$, betas of 0.9 and 0.999, weight decay of 0. 
    For experiments with the MNIST dataset, the data augmentation applied included a random crop transformation with padding 4, followed by a normalization transformation. 
    For FashionMNIST, a similar data augmentation was used, with the addition of a random horizontal flip.
    Below, we report the specific settings used for particular models.
    
    \subsubsection{Experiments with SmallConv and SmallConvL}
    For experiments with the SmallConv and SmallConvL models, we used light data augmentation for CIFAR10, CIFAR100, and IMAGENETTE. 
    For CIFAR10 and CIFAR100, only a random horizontal flip was applied. 
    For IMAGENETTE, the images were resized to 132x132 pixels and then randomly cropped to 128x128 pixels, followed by a random horizontal flip. 
    The models were trained for 100 epochs for the experiments reported in Table 1 and Table 2.
    
    \subsubsection{Experiments with VGG8}
    We used more extensive data augmentation for experiments with CIFAR10, CIFAR100, IMAGENETTE, and TinyIMAGENET. 
    The data augmentation consisted of a random crop, followed by a random horizontal flip, then a normalization layer, and a random erasing \cite{Zhong2017RandomAugmentation} with a probability of 0.2. 
    When VGG8 was trained on MNIST and FashionMNIST, the model was trained for 100 epochs. 
    For the other datasets, the model was trained for 300 epochs and dropout layers with a probability of 0.2 were used after each ConvBlock.
    
    \subsubsection{Experiments with MobileNetV1}
    For the experiments with the Visual Wake Words (VWW) dataset, the training images were resized and randomly cropped to a size of 128x128 pixels, followed by normalization. 
    The model was trained for 500 epochs for the experiments reported in Table 3.
        
    \subsection{Experimental Compute Resources}\label{appendix:compute_resources}
    All experiments were conducted on a shared internal Linux server equipped with an AMD EPYC 7502 32-Core Processor, 504 GB of RAM, and four NVIDIA A40 GPUs, each with 48 GB of GDDR6 memory.
    Additionally, code was implemented using Python 3.9 and PyTorch 2.2.1 with CUDA 11.8.

\end{document}